\documentclass{article}
\usepackage{spconf,amsmath,graphicx}

\usepackage{blindtext}

\usepackage{hyperref}
\usepackage{color}
\hypersetup{
    colorlinks=true,
    linkcolor=blue,
    filecolor=magenta,      
    urlcolor=cyan,
    pdftitle={Overleaf Example},
    pdfpagemode=FullScreen,
    }
    
\title{Explainable AI (XAI) in Biomedical Signal and Image Processing: Promises and Challenges}
\name{Guang Yang$^1$\thanks{G.Y. is supported by the ERC IMI (101005122), the H2020 (952172), the MRC (MC/PC/21013), and the UKRI Future Leaders Fellowship (MR/V023799/1). Email: g.yang@imperial.ac.uk}, Arvind Rao$^2$\thanks{A.R is supported by NIH grants R37CA214955-01A1, a Research Scholar Grant from the ACS RSG-005-016, and MIDAS/MICDE funding from the University of Michigan Ann Arbor. Email: ukarvind@umich.edu}, Christine Fernandez-Maloigne$^3$, Vince Calhoun$^4$\thanks{V.C. was partly supported by the National Science Foundation 2112455}, Gloria Menegaz$^5$\thanks{G.M. was partly supported by Fondazione CariVerona (Bando Ricerca Scientifica di Eccellenza 2018, EDIPO project - reference number 2018.0855.2019)}}
\address{$^1$ National Heart and Lung Institute, Imperial College London, UK ~\\ $^2$ Dept. of Computational Medicine and Bioinformatics, University of Michigan, Ann Arbor, USA ~\\$^3$ I3M Laboratory, University of Poitiers and CNRS, France ~\\$^4$ TReNDS\ Center, Georgia State, Georgia Tech, Emory, Atlanta, GA, USA ~\\$^5$ Dept. of Computer Science, University of Verona, Italy }
\begin{document}

%
\maketitle

\begin{abstract}
Artificial intelligence has become pervasive across disciplines and fields, and biomedical image and signal processing is no exception. The growing and widespread interest on the topic has triggered a vast research activity that is reflected in an exponential research effort. Through study of massive and diverse biomedical data, machine and deep learning models have revolutionized various tasks such as modeling, segmentation, registration, classification and synthesis, outperforming traditional techniques. However, the difficulty in translating the results into biologically/clinically interpretable information is preventing their full exploitation in the field. Explainable AI (XAI) attempts to fill this translational gap by providing means to make the models interpretable and providing explanations. Different solutions have been proposed so far and are gaining increasing interest from the community.
This paper aims at providing an overview on XAI in biomedical data processing and points to an upcoming Special Issue on Deep Learning in Biomedical Image and Signal Processing of the IEEE Signal Processing Magazine that is going to appear in March 2022. 
\end{abstract}
\begin{keywords}
Explainable AI, XAI,  Artificial Intelligence, Machine Learning, Deep Learning, Biomedical Data
\end{keywords}
\section{Introduction}
Recent years have witnessed an increased interest in  artificial intelligence (AI), in particular deep learning (DL) and machine learning (ML) techniques, as a result of recent breakthroughs. DL and ML have become pervasive in many fields including biomedical signal and image processing where multimodal, multidimensional, multiparametric datasets need to be jointly processed, posing a clear challenge to traditional analysis methods.
However, while AI algorithms are designed to assist users with their regular tasks, there still remain challenges with acceptance. Users frequently express lack of trust with respect to outcomes of such methods, {\em de-facto} impeding or slowing-down their exploitation with obvious consequences. This motivates the need for methods to equip them with those attributes making them understandable, trustable and prone to be validated through the matching of the outcomes with known priors or association studies.
All this puts eXplainable AI (XAI) at the core of the AI-driven shift in the biomedical field, as one of the main challenges leading to the next generation of human-centric intelligent systems.
This paper aims at providing an overview on the main XAI contributions in the biomedical field, across methods and modalities, either by analyzing and understanding algorithm components or by unraveling algorithm-data interactions. This will provide some guidance to the entrants while highlighting the main issues and challenges to be overcome in the next few years. For doing this, some showcases will be considered, providing examples of the application of XAI methods to concrete problems in biomedical sciences. An extended overview is provided in the 
 \href{http://signalprocessingsociety.org/blog/ieee-spm-special-issue-deep-learning-biological-image-and-signal-processing}{Special Issue on Deep Learning in Biomedical Image and Signal Processing} of the IEEE Signal Processing Magazine that was recently published (Feb. 2022).
This paper is organized as follows. Section \ref{xai} provides a short overview on XAI methods, relating to recently published survey papers. Section \ref{biomed} discusses some issues that are specific to the biomedical field. Section \ref{case_studies} presents some showcases and Section \ref{challenges} highlights some of the major challenges to be faced in order to overcome the main bottlenecks contrasting the full adoption of such methods.

\section{eXplainable AI: a brief survey}
\label{xai}
XAI recently emerged as one of the hottest topics for understanding ”the why and how” of the outcomes of ML/DL algorithms. A comprehensive overview can be found in \cite{linardatos2021explainable,holzinger2019causability,arrieta2020explainable,yang2022unbox}, where the main concepts are illustrated, a taxonomy is proposed and the main state-of-the-art approaches are summarized and assigned to the respective category.
Before presenting the main methods, it is important to devote a few words to the terminology.
Following \cite{linardatos2021explainable}, \textit{interpretability} is connected with the human intuition behind the outputs of a model, supporting the understanding of cause-and-effect relationships within the system input and output. This in turn points to \textit{causability}. Instead, following \cite{linardatos2021explainable}, \textit{explainability} would be associated with the decoding of the internal logic and mechanisms of a ML system, hence it is not necessarily related to human understanding. 
Therefore, interpretability does not axiomatically entail explainability and vice versa, following \cite{linardatos2021explainable}.
In order for a system to be interpreted, \emph{explanations} must be derived and the properties making an explanation effective to humans need to be defined.
Though a common agreement on the terminology and taxonomy is still to be reached, among the most used attributes are \cite{holzinger2019causability} \textit{post-hoc} or \textit{ante-hoc}. While the latter are explainable models that also hold the interpretability property by construction, the former
are "black-box" whose interpretation relies on post-hoc methods that in turn can be either model-agnostic or model-specific, as well as local or global. We refer to \cite{linardatos2021explainable} for a comprehensive review.
Widely used post-hoc interpretability methods for black-box models rely on feature probing such as perturbation-based methods. Among these, the SHapley Additive exPlanation (SHAP) \cite{lundberg2017unified} and the Local Interpretable Model-Agnostic Explanations (LIME) \cite{ribeiro2016model} ones recently gained large popularity. 
Moving to DL, many methods have been devised, mostly relying on back propagating gradients. Among these are SmoothGrad, Class Activation Mapping (CAM) and Grad-CAM. Other popular methods are DeepLIFT (Deep Learning Important FeaTures) \cite{shrikumar2017learning}, and Layerwise Relevance Propagation \cite{Bohle_lrp}. These generate heatmaps highlighting the image elements that mostly contributed to the outcomes.

\section{XAI in the biomedical field}
\label{biomed}
This Section provides some hints on the main requirements originating from the specificity of data and explanation in the biomedical field, that will then be exemplified in three showcases in Section \ref{case_studies}. 
\subsection{Biomedical Data}
Biomedical data include a wide range of modalities and covers several orders of magnitude in both space and time. 
In addition to being highly heterogeneous, biomedical data are noisy, suffering from missing data, and strongly dependent on the acquisition devices, so that a path must be traced in the space of algorithmic solutions allowing to deal with such a heterogeneous environment. 
Indeed, from tiny molecules to omic data (e.g., genomic, proteomic, transcriptomic, metabolomic), imaging data, clinical data, and electronic medical records, biomedical data  span a very large space. Data quality may also be highly variable, and data formats can range from analogue to digital to text, with sophisticated linked structures such as sequences, trees, and other graphs that can vary in size. 
Improvements in sensors and other devices, as well as computers, databases, along with the advent of innovative high-volume data generation procedures like high-throughput sequencing, functional and diffusion weighted magnetic resonance imaging (MRI), have resulted in a data avalanche that requires {\em ad-hoc} processing solutions. Even in the era of big data, however, the data environment remains quite volatile in terms of the amount of data accessibility. 
%
While the most natural answer to deal with such massive heterogeneous data is DL, the challenge is to exploit these methods even when data is scarce, as it is often the case for local data. 
Several techniques, ranging from regularisation methods, including some relatively new forms of regularisation such as dropout, to early stopping, semi-supervised approaches that try to leverage both labelled and unlabeled data, and data augmentation methods, where datasets are expanded using natural or artificially generated data (for example, by adding appropriate noise or applying appropriate group transformations) have been proposed. This adds levels of complexity due to data manipulation, that is particularly delicate in the biomedical field where congruence with biological, physiological or other kind of prior knowledge must be granted. 
The analysis of the different challenges posed by biomedical data and potential solutions is out of the scope of this contribution. Here we aim at providing a view on one aspect, that is interpretability, and to illustrate the concepts through some examples. In summary, biomedical data pose a great challenge to XAI, calling for solutions that also take uncertainty into account, and thus validation of the outcomes of the XAI methods, which is still scarcely explored.

\subsection{Explanations: basic requirements and validation}
In the biomedical field interpretability is an important factor, while validation is an essential issue \cite{kohoutova2020toward}. At a minimum, models must be understandable by humans, trustable and robust. Once the explanations are extracted, these need to be validated. The validation phase is important in the biomedical field, and concerns different aspects. Explanations must be plausible in the biomedical sense to be reliable, and stable across implementations (methods and architectures) to generate trust. In addition, they should not be affected by the presence of confounders, and must prove statistical significance. 
Among the most exploited methods are those relying on heatmaps revealing which parts of the input played a relevant role in determining the output.
An example of a comprehensive solution was proposed in \cite{Cruciani2021}, where LRP was used to interpret the results of a patient stratification task aiming at distinguishing two multiple sclerosis phenotypes (progressive versus relapsing-remissing) and validation included the assessment of the role of confounds on the classification outcomes as well as association studies of LPR maps with microstructural descriptors that were previously shown to be significantly different in the two phenotypes of disease.
Another example is \cite{cfm1} where the objective was to contrast feature-based and example-based explainable AI techniques using the Heart Disease Dataset from UCI  (https://www.kaggle.com/cherngs/ heart- disease- cleveland- uci) to highlight the respective pros and cons.
A different perspective is given in \cite{cfm2}, suggesting that explainability should be explicated as ‘effective contestability’. Taking a patient-centric approach the authors argue that patients should be able to contest the diagnoses of AI diagnostic systems, and that effective contesting of patient-relevant aspect of AI diagnoses requires the availability of different types of information about the AI system’s use of data, the system’s potential biases and the system performance.

The next Section highlights some of these issues in more details through the illustration of three showcases relying on different types of data and algorithms.
%
\section{Case Studies}
\label{case_studies}
\subsection{Neuroimaging: Brain Aging}
The study of brain aging has recently gained attention in the scientific community since developing accurate biomarkers for Brain Age (BA) relying on neuroimaging data would open new perspectives in different domains, allowing to disentangle age-related from disease-specific changes and to track the disease progression at the single-subject level. Based on a given set of predictors, the "brain age delta" representing the difference between the actual and the estimated age is inferred. For this task, ante-hoc models have heavily been exploited including linear models, latent variable models, such as PCA and ICA \cite{smith2020brain,cole2020multimodality}, while post-hoc ones are gaining importance. Among these are permutation- and perturbation-based \cite{kolbeinsson2020accelerated,kaufmann2019common} feature importance, and saliency maps.
Among those using gradient based methods, \cite{levakov2020deep} exploited SmoothGrad to produce explanation maps for a CNN model and \cite{wang2019gray,feng2020estimating} obtained Grad-CAM attention maps from T1w-based CNN showing the relative importance of different regions for BA prediction. 
The second also applied ablation analyses methods focusing on part of the input data. 
Deconfounding and bias correction are most often applied, and association studies are performed between this index and other characteristics measured in the same individuals, related for example to cognitive, physical, lifestyle and genetic determinants.
In summary, the key aspects of XAI adoption in BA estimation are the following. First, despite wide use of explainable models such as linear regression or regularized regression, the respective explanations are still not pervasively derived. Most recently, visualization of the model coefficients in the latent space is being increasingly exploited, and permutation- and perturbation-based feature importance methods play a central role in extracting explanations. Saliency maps, even if still highly unexplored, are increasingly used following the spread of DL models and mostly rely on gradient-based methods \cite{Boscolo-2022}.
\subsection{COVID-19 Diagnosis}
CT scans are useful for determining the severity of an infection and can also be utilized to detect COVID-19 carriers who are symptomatic or asymptomatic. With an increase in the total number of COVID-19 patients, radiologists are under more pressure to manually evaluate CT images. As a result, an automatic 3D CT scan identification tool is in great demand, as manual analysis takes time for radiologists, and their weariness might lead to mistakes. However, the look of CT scans can fluctuate greatly depending on the technical specifications of CT scanners used in different institutions, causing many automatic image identification systems to fail. As a result, the multi-domain shift problem for multi-center and multi-scanner research is not straightforward, but it is also vital for reliable detection and repeatable and objective diagnosis and prognosis. Ye et al. \cite{ye2022robust} introduced the coronavirus information fusion and diagnostic network (CIFD-Net), a COVID-19 CT scan identification model that can easily tackle the multi-domain shift problem using a new robust weakly supervised learning approach. CNN's prediction process is kept in a black box. Several strategies have been presented to explain how CNN produces predictions and achieves exceptional localization abilities without the need for localization maps, e.g., CAM. The updated activation mapping in Ye et al. \cite{ye2022robust} can properly show the relevance of activation from CT scans and find the infected regions of COVID-19 patients, resulting in predictable and understandable outcomes. The region with a higher activation score suggests that the activation played a larger role in the prediction. For radiologists, the updated activation mapping might provide additional diagnostic information.
\subsection{XAI in multiplex immunofluorescence (mIF) images for Cancer Immunotherapy}
The advent of cancer immunotherapy has necessitated the use and employment of in-situ, tissue based spatial profiling tools (like spatial transcriptomics, CODEX/MIBI, spatial proteomics etc, (\cite{10.3389/fimmu.2022.847582}, \cite{lewis_asselin-labat_nguyen_berthelet_tan_wimmer_merino_rogers_naik_2021}) to characterize the cellular milieu and tumor microenvironment underlying cancer tissue. Specifically, the spatial patterns of cell-cell organization permit the characterization and inference of spatial biomarkers \cite{10.3389/fimmu.2022.847582} that might be prognostic of disease state or prognosis. \cite{10.3389/fimmu.2021} proposed a framework, Cell Graph ATtention Network (CGAT) that uses deep learning graph architectures to analyze and interpret mIF images of tissue. While the GCN framework is used for classification of disease grade, the network architecture includes an attention layer that identifies particular nodes that are most relevant to the prediction task. The examination of the CGAT framework to classify pancreatic disease obtained favorable performance in addition to identifying graph nodes (cells in the tissue) most relevant to prediction. Cells at the interface of tumor and stromal regions (i.e invasive margin) were identified as predictive thereby adding interpretability and explainability to the inference task. Such tools will hopefully enable the development of integrated user-interfaces that couple high performance predictive modeling and their principled interpretation by domain experts.
\section{Discussion and Future Perspectives}
\label{challenges}
XAI is a set of processes and methods that allow human users to understand and trust the results and conclusions created by machine learning algorithms. XAI is used to describe an AI model, its expected impact, and its potential biases. It helps characterize the accuracy, fairness, transparency, and outcomes of models in AI-assisted decision making. 
However, injecting XAI in biomedical signal and image processing is far from trivial. 
First, a clear definition of the concepts and definitions is still missing. This is a difficult interdisciplinary issue involving philosophical, psychological and ethical aspects. Then, once the concepts and definitions are given, attributes must be defined as well as measures, or {\em proxies} to assess them. A first attempt to classify the attributes for explanations can be found in \cite{doshi2017towards}, identifying three types of attributes to the validation of an interpretability method that is \emph{application-grounded, human-grounded and functionality-grounded}. While the first two take the user in the loop, the last one relies on objective assessment through proxies. An extended taxonomy can be found in \cite{nauta2022anecdotal} where twelve \emph{explanation quality properties}, grouped by their most prominent dimension, that is Content, Presentation or User, are proposed.
Given the attributes and the measures, validation is the critical step. As mentioned before, besides being interpretable, the explanations must be reliable and robust, which imply association studies with eloquent disease indices and assessment of the concordance with prior knowledge, when available. In addition, the lack of ground truth adds a level of complexity, that cannot be overcome and requires {\em ad-hoc } solutions.
To generate trust, the outcomes should be implementation-independent, and the uncertainty in the measures needs also to be assessed. This is model-specific and comes from many sources, including the quality and quantity of the data, which makes it a very challenging problem to assess the concordance across methods and express it in quantifiable and understandable terms. 
More broadly, the development of ML/DL approaches that are informed by "domain priors" in biomedical science might serve as useful ways to engineer interpretabilty and explainability into these systems. 
%
%
A clear trade-off has also emerged between performance and explainability, leading to the adoption of solutions that prioritize one or the other based on the scientific question. 
%
Learning biases introduced by DL methods used in biomedical data science might hinder AI approaches from delivering the bare minimum of interpretations. The learning bias problem refers to the fact that AI findings may be skewed or even incorrect \cite{han2015diagnostic}. These can be produced by inappropriate interactions between particular AI algorithms and a specific type of data, incorrect parameter setup or tuning, unbalanced data, or other more intricate difficulties, that biomedical data scientists may not be able to detect readily. Because of the artefacts in AI models, learning bias is technically a learning security concern that creates unmanageable effects. 
%
In recent research efforts, rule-based learning, learning process visualisation, knowledge-based data representation, human-centred AI-model evaluation and other techniques have been used to improve AI explainability. The approaches will surely aid in the development of explainable AI in biological data science. However, additional attention and effort in biomedical data science research may be required to overcome the hurdles and produce explainable and efficient AI systems. 
Validation of methods and explanations through objective assessment of the required attributes such as stability (across methods and data), plausibility and uncertainty, just to mention some of the main ones, is the key to empower XAI in this field, and is one of the hottest topics being investigated.

In summary, the goal of this paper was to rise awareness on specific facets of XAI in the biomedical field, providing some hints for boosting AI exploitation and trust. 


\bibliographystyle{IEEEbib}
\bibliography{refs}

\begin{thebibliography}{10}

\bibitem{yang2022unbox}
G.~Yang, Q.~Ye, and J.~Xia,
\newblock ``Unbox the black-box for the medical explainable ai via multi-modal
  and multi-centre data fusion: A mini-review, two showcases and beyond,''
\newblock {\em Information Fusion}, vol. 77, pp. 29--52, 2022.

\bibitem{ye2021explainable}
Q.~Ye, J.~Xia, and G.~Yang,
\newblock ``Explainable ai for covid-19 ct classifiers: an initial comparison
  study,''
\newblock in {\em 2021 IEEE CBMS}. IEEE, 2021, pp. 521--526.

\bibitem{linardatos2021explainable}
P~Linardatos, V.~Papastefanopoulos, and S.~Kotsiantis,
\newblock ``Explainable ai: A review of machine learning interpretability
  methods,''
\newblock {\em Entropy}, vol. 23, no. 1, pp. 18, 2021.

\bibitem{holzinger2019causability}
A.~Holzinger, G.~Langs, H.~Denk, K.~Zatloukal, and H.~M{\"u}ller,
\newblock ``Causability and explainability of artificial intelligence in
  medicine,''
\newblock {\em Wiley Interdisciplinary Reviews: Data Mining and Knowledge
  Discovery}, vol. 9, no. 4, pp. e1312, 2019.

\bibitem{arrieta2020explainable}
A.~B. Arrieta, N.~D{\'\i}az-Rodr{\'\i}guez, J.~Del~Ser, Bennetot, et~al.,
\newblock ``Explainable artificial intelligence (xai): Concepts, taxonomies,
  opportunities and challenges toward responsible {AI},''
\newblock {\em Information fusion}, vol. 58, pp. 82--115, 2020.

\bibitem{lundberg2017unified}
S.M. Lundberg and S.~Lee,
\newblock ``A unified approach to interpreting model predictions,''
\newblock {\em Advances in neural information processing systems}, vol. 30,
  2017.

\bibitem{ribeiro2016model}
M.T. Ribeiro, S.~Singh, and C.~Guestrin,
\newblock ``Model-agnostic interpretability of machine learning,'' 2016.

\bibitem{shrikumar2017learning}
A.~Shrikumar, P.~Greenside, and A.~Kundaje,
\newblock ``Learning important features through propagating activation
  differences,''
\newblock in {\em International conference on machine learning}. PMLR, 2017,
  pp. 3145--3153.

\bibitem{Bohle_lrp}
M~Böhle, F.~Eitel, M.~Weygandt, and K.~Ritter,
\newblock ``Layer-wise relevance propagation for explaining deep neural network
  decisions in mri-based alzheimer's disease classification,''
\newblock {\em Frontiers in Aging Neuroscience}, vol. 11, pp. 194, 2019.

\bibitem{kohoutova2020toward}
L.~Kohoutov{\'a}, J.~Heo, S.~Cha, S.~Lee, T.~Moon, T.D. Wager, and C.-W. Woo,
\newblock ``Toward a unified framework for interpreting machine-learning models
  in neuroimaging,''
\newblock {\em Nature protocols}, vol. 15, no. 4, pp. 1399--1435, 2020.

\bibitem{Cruciani2021}
F.~Cruciani, L.~Brusini, M.~Zucchelli, G.~Retuci~Pinheiro, F.~Setti,
  I.~Boscolo~Galazzo, R.~Deriche, L.~Rittner, M.~Calabrese, and G.~Menegaz,
\newblock ``Interpretable deep learning as a mean for decrypting disease
  signature in multiple sclerosis,''
\newblock {\em Journal of Neural Engineering}, vol. 18, no. 4, pp. 0460a6,
  2021.

\bibitem{cfm1}
D.~Dave, H.~Naik, S.~Singhal, and P.~Patel,
\newblock ``Explainable ai meets healthcare: A study on heart disease
  dataset,'' 2020.

\bibitem{cfm2}
T.~Ploug and S.~Holm,
\newblock ``The four dimensions of contestable ai diagnostics - a
  patient-centric approach to explainable ai,''
\newblock {\em Artificial Intelligence in Medicine}, vol. 107, July 2020.

\bibitem{smith2020brain}
S.M. Smith, L.T. Elliott, F.~Alfaro-Almagro, et~al.,
\newblock ``Brain aging comprises many modes of structural and functional
  change with distinct genetic and biophysical associations,''
\newblock {\em Elife}, vol. 9, pp. e52677, 2020.

\bibitem{cole2020multimodality}
J.H. Cole,
\newblock ``Multimodality neuroimaging brain-age in uk biobank: relationship to
  biomedical, lifestyle, and cognitive factors,'' 2020.

\bibitem{kolbeinsson2020accelerated}
A.~Kolbeinsson, S.~Filippi, Y.~Panagakis, P.M. Matthews, P.~Elliott, et~al.,
\newblock ``Accelerated mri-predicted brain ageing and its associations with
  cardiometabolic and brain disorders,''
\newblock {\em Scientific Reports}, vol. 10, no. 1, pp. 1--9, 2020.

\bibitem{kaufmann2019common}
T.~Kaufmann, D.~van~der Meer, N.T. Doan, E.~Schwarz, et~al.,
\newblock ``Common brain disorders are associated with heritable patterns of
  apparent aging of the brain,''
\newblock {\em Nature neuroscience}, vol. 22, no. 10, pp. 1617--1623, 2019.

\bibitem{levakov2020deep}
G.~Levakov, G.~Rosenthal, I.~Shelef, et~al.,
\newblock ``From a deep learning model back to the brain---identifying regional
  predictors and their relation to aging,''
\newblock {\em Human brain mapping}, vol. 41, no. 12, pp. 3235--3252, 2020.

\bibitem{wang2019gray}
J.~Wang, M.J. Knol, A.~Tiulpin, et~al.,
\newblock ``Gray matter age prediction as a biomarker for risk of dementia,''
\newblock {\em Proceedings of the National Academy of Sciences}, vol. 116, no.
  42, pp. 21213--21218, 2019.

\bibitem{feng2020estimating}
X.~Feng, Z.C. Lipton, J.~Yang, S.A. Small, and F.A. Provenzano,
\newblock ``Estimating brain age based on a uniform healthy population with
  deep learning and structural magnetic resonance imaging,''
\newblock {\em Neurobiology of aging}, vol. 91, pp. 15--25, 2020.

\bibitem{Boscolo-2022}
I.~Boscolo~Galazzo, F.~Cruciani, L.~Brusini, A.~Salih, P.~Radeva, S.F. Storti,
  and G.~Menegaz,
\newblock ``Explainable deep learning for mri aging brainprints: grounds and
  challenges,''
\newblock {\em IEEE-SPM, Special Issue on Deep Learning in Biological Signal
  and Image Processing}, 2022.

\bibitem{ye2022robust}
Q.~Ye, Y.~Gao, W.~Ding, Z.~Niu, C.~Wang, et~al.,
\newblock ``Robust weakly supervised learning for covid-19 recognition using
  multi-center ct images,''
\newblock {\em Applied Soft Computing}, vol. 116, pp. 108291, 2022.

\bibitem{10.3389/fimmu.2022.847582}
D.~Phillips, S.~J. Rodig, and S.~Jiang,
\newblock ``Editorial: Defining the spatial organization of immune responses to
  cancer and viruses in situ,''
\newblock {\em Frontiers in Immunology}, vol. 13, 2022.

\bibitem{lewis_asselin-labat_nguyen_berthelet_tan_wimmer_merino_rogers_naik_20%
21}
S.M. Lewis, M.L. Asselin-Labat, Q.~Nguyen, J.~Berthelet, X.~Tan, V.C. Wimmer,
  D.~Merino, K.L. Rogers, and S.H. Naik,
\newblock ``Spatial omics and multiplexed imaging to explore cancer biology,''
\newblock {\em Nature Methods}, vol. 18, no. 9, pp. 997–1012, 2021.

\bibitem{10.3389/fimmu.2021}
M.~Baranwal, S.~Krishnan, M.~Oneka, T.~Frankel, and A.~Rao,
\newblock ``Cgat: Cell graph attention network for grading of pancreatic
  disease histology images,''
\newblock {\em Frontiers in Immunology}, vol. 12, 2021.

\bibitem{doshi2017towards}
Finale Doshi-Velez and Been Kim,
\newblock ``Towards a rigorous science of interpretable machine learning,''
\newblock {\em arXiv preprint arXiv:1702.08608}, 2017.

\bibitem{nauta2022anecdotal}
J.~Nauta, M.~Trienes, P.~Shreyasi, et~al.,
\newblock ``From anecdotal evidence to quantitative evaluation methods: A
  systematic review on evaluating explainable ai,'' 2022.

\bibitem{han2015diagnostic}
H.~Han,
\newblock ``Diagnostic biases in translational bioinformatics,''
\newblock {\em BMC Medical Genomics}, vol. 8, no. 1, pp. 1--17, 2015.

\end{thebibliography}

\end{document}